\documentclass[conference]{IEEEtran}
\usepackage[utf8]{inputenc}
\usepackage[T1]{fontenc}
\usepackage[english]{babel}
\usepackage{fancyhdr}
\usepackage{a4wide}

\usepackage{amsmath}
\usepackage{amssymb} 
\usepackage{amsfonts}

\usepackage{graphicx}

\usepackage{fancybox}
\usepackage{listings}

\usepackage[hidelinks]{hyperref}
\usepackage{calc}
\usepackage{float}

\usepackage[official]{eurosym}
\usepackage[final]{pdfpages}
\usepackage{xspace}
\usepackage{caption}
\usepackage{subcaption}
\usepackage{mathtools}

\usepackage{enumitem}

\usepackage{amsmath}
\usepackage{amssymb}
\usepackage{mathrsfs} 
\usepackage{amsthm}
\usepackage{amssymb} 
\usepackage{amsfonts}
\usepackage{mathabx}
\usepackage{algorithm2e}

\usepackage{verbatim} 
\usepackage{mathptmx} 
\usepackage{dsfont}
\usepackage{graphicx} 
\usepackage{color} 
\usepackage[super]{nth}
\definecolor{bblack}{cmyk}{0,0,0,1} 
\usepackage{listings}
\definecolor{dkgreen}{rgb}{0,0.4,0}
\definecolor{gray}{rgb}{0.5,0.5,0.5}
\definecolor{mauve}{rgb}{0.58,0,0.82}
\definecolor{dkyellow}{cmyk}{0, 0, 0.2, 0}
\usepackage{graphicx}
\usepackage{algorithmic}

\usepackage{todonotes}

\title{Introducing machine learning \\for power system operation support}
\author{Benjamin DONNOT${^\ddag} ^{+\dagger}$, Isabelle GUYON$^{*+\ddag}$, Marc SCHOENAUER$^+\ddag$,\\ Patrick PANCIATICI$^\dagger$, Antoine MAROT$^\dagger$ \\
{\centering *UPSud Paris-Saclay, +INRIA $^\ddag$LRI, Laboratoire de Recherche en Informatique} \\ 
{\centering$^\dagger$RTE R\&D}
} 
\date{January 2017}

\usepackage{ulem}

\newcommand*{\MyDef}{\mathrm{def}}
\newcommand*{\eqdefU}{\ensuremath{\mathop{\overset{\MyDef}{=}}}}
\newcommand*{\eqdef}{\mathop{\overset{\MyDef}{\resizebox{\widthof{\eqdefU}}{\heightof{=}}{=}}}}

\newcommand{\abs}[1]{\left\lvert #1 \right\rvert}

\begin{document}

\maketitle
\begin{abstract}
We address the problem of assisting human dispatchers in operating power grids in today's changing context using {\it machine learning}, with the aim of increasing security and reducing costs. Power networks are highly regulated systems, which at all times must meet varying demands of electricity with a complex production system, including conventional power plants, less predictable renewable energies (such as wind or solar power), and the possibility of buying/selling electricity on the international market with more and more actors involved at a European scale. This problem is becoming ever more challenging in an aging network infrastructure.
One of the primary goals of dispatchers is to protect equipment (e.g. avoid that transmission lines overheat) with few degrees of freedom: we are considering in this paper solely modifications in network topology, i.e. re-configuring the way in which lines, transformers, productions and loads are connected in sub-stations.  
Using years of historical data collected by the French Transmission Service Operator (TSO) ``R\'eseau de Transport d'Electricit\'e" (RTE), we develop novel machine learning techniques (drawing on ``deep learning") to mimic human decisions to devise ``remedial actions" to prevent any line to violate power flow limits (so-called  "thermal limits"). The proposed technique is hybrid. It does not rely purely on machine learning: every action will be tested with actual simulators before being proposed to the dispatchers or implemented on the grid.


\end{abstract}


\section{Introduction \label{seq:intro}}
Electricity is a commodity that consumers take for granted and, while governments relaying public opinion (rightfully) request that renewable energies be used increasingly, little is known about what this entails behind the scenes in additional complexity for the Transmission Service Operators (TSOs) to operate the power grid in security. Indeed, renewable energies such as wind and solar power are less predictable than conventional power sources (mainly thermal power plants).
In cases of contingency, which may be weather-related (e.g. decreased production because of less wind or sun or line failure due to meteorological conditions) operators (a.k.a. dispatchers) must act quickly to protect equipment to meet all ``security criteria" (for example to avoid that lines get overloaded). \textit{Remedial actions} they take in such situations may include among others (1) modifications of the network {\bf topology} to re-direct power flows, (2) modification of productions or consumptions ({\bf re-dispatching}). By far the least costly and preferred of these options is the first one, and it will be the only one considered in this paper.
A network is considered to be operated in ``security" (i.e. in a secure state) if it is outside a zone of ``constraints", which includes that power flowing in every line does not exceed given limits.
The dispatchers must avoid ever getting in a critical situation, which may lead to a cascade of failures (circuit breakers opening lines automatically to protect equipment, thus putting more and more load on fewer and fewer lines), ultimately leading to a blackout. To that end, it is standard practice to operate the grid in real time with the so-called ``N-1 criterion": 
this is a preventive measure requiring that at all times the network would remain in a safe state even if one component (productions, lines, transformers, etc.) would be disconnected.

In choosing proper remedial actions, the dispatchers are facing various trade-offs.
Remedial actions must eliminate the problem they were designed to address, but also must avoid creating new problems elsewhere on the grid.
Today, the complex task of dispatchers, which are highly trained engineers, consists in analyzing situations, proposing remedial actions, and checking prospectively their effect using sophisticated (but slow) high-end simulators, which allow them to investigate only a few options. Our goal is to  assist the dispatchers by suggesting them with quality candidate remedial actions, obtained by synthesizing several years of historical decisions made in various situations into a powerful predictive machine learning models, built upon earlier work~\cite{wehenkel2012automatic}.\\

The main contributions of this paper are: (1) To address a large scale industrial project with potentially high financial impact using real historical data and a large-scale simulator (deployed in real operations) from the company RTE; (2) To cast the problem in a mathematical setting amenable to machine learning studies; (3) To devise a methodology to extract from historical data and simulations a dataset usable for training and testing in a supervised machine learning setting; (4) To suggest and study machine learning architectures, which automatically generate candidate remedial actions, which could be validated with more extensive power system simulations.
The paper is organized as follows:
Section \ref{seq:pb} formalizes the problem. 
Section \ref{seq:method} describes the proposed methodology. 
Section \ref{sec:results} outlines initial results. 
Section \ref{seq:inte} presents a possible integration into today operational processes. Finally, section~\ref{sec:conclusion} provides conclusions and outlooks. \\


\section{Formalization of the problem. \label{seq:pb}}
In this section, we formalize daily real-time tasks of dispatchers as a formal realistic (yet simplified) optimization problem, amenable to mathematical studies. Our setting is inspired by the analysis found in reference~\cite{kundur2012power}\\ 

Suppose that we are studying a powergrid at a given time $t$ (either the current time for a real-time study, or some time in the near future for a forecast study). Let:
\begin{itemize}
\item $\mathcal{R}_t$ be the set of all feasible re-dispatching actions possible for time $t$; 
\item and $\mathcal{T}_t$ the set of all feasible topological actions for time $t$
\end{itemize}
\noindent known at the time of the study. \\

Let us then assume that we are given a cost function $R$ (resp. $T$), that assigns some cost to any re-dispatching action $\rho \in \mathcal{R}_t$ (resp. topological action $\tau \in \mathcal{T}_t$). For instance, the cost of a redispatching action can be the money paid by the TSO to the producers. The cost of a topological action can include the aging of the breakers, the probability of failure etc.


We further assume that decisions performed by dispatchers made for the sake of security of the grid are optimally efficient, given available information. They implicitly solve an optimization problem consisting in minimizing the cost of their actions to meet a security measure $\mathbb{S}$. This can be formalized with the equation \ref{eq:pbdisp}:
\begin{equation}
\label{eq:pbdisp}
\begin{aligned}
\underset{(\rho \in \mathcal{R}_t, \tau \in \mathcal{T}_t) }{ \text{minimize}} &R(\rho) + T(\tau) \\
 \text{subject to}  \\
 & \mathbb{S}(\text{grid}_t\odot\{\rho,\tau \})
\end{aligned}
\end{equation}
\noindent where $\mathbb{S}$ denotes the function stating whether a powergrid is in a secure state. More formally, $\mathbb{S}$ should be a function taking a grid as input, and returning a list of security issues (for example if the grid is secure according to $\mathbb{S}$, the result should be the empty set $\emptyset$). We also denote by $\text{grid}_t$ the state of the grid at time $t$. The operator $\odot$ must be understood as applying a set of actions on a given grid: "$\text{grid}_t\odot\{\rho,\tau \}$" should be though as \textit{The grid resulting of the application of actions $\rho$ and $\tau$ on the network $\text{grid}_t$}. 

This problem can be very complex to solve. For instance, it mixes continuous variables (such as redispatching) and integer variables (for example the topology or the maximum values allowed for productions).
The number of variables involved is also quite important. France alone count around $3~000$ productions and RTE can act on more than $30~000$ breakers. 
Solving this problem "as is" requires to do some hypothesis on the costs functions and on the type of constraints of the problem \ref{eq:pbdisp} for example to formulate as a Mixed-Integer Linear Program for which there exist some suitable solvers.

In this paper, we propose a new methodology, based on learning of remedial actions taken by operators. Indeed, learning from human actions has some advantages:
\begin{itemize}
\item It will improve the acceptance of the algorithm for dispatcher:
\begin{itemize}
\item the proposed actions come from what they have already done in the past;
\item they can use the same tools they use today to check the validity of the results proposed.
\end{itemize}
\item It can indirectly model other security issues ignored by $\mathbb{S}$. For example, dispatchers may know that a given breaker is in bad shape. So, they rarely actuated it. This can be taken into account by a learning strategy but may not be as easily digested using optimization tools (such constraints may be difficult to express, or difficult to centralized in one unique Information System Database).
\item It can help sharing knowledge between dispatchers, and capitalizing on the best action taken.
\end{itemize}

\section{Proposed methodology\label{seq:method}}
In this section, we address the problem of finding curative/remedial actions to protect the power grid with a novel methodology based on machine learning. Our methodology is inspired by the game playing literature and in particular the very successful AlphaGO machine learning program\cite{silver_mastering_2016} developed by Google Deepmind to tackle the ancient game of Go. We detail in this section the first step of the methodology concerning ``imitation learning", i.e. training a learning machine to imitate decisions made by experts (expert players for Go and professional dispatchers for power grids). Improvements gained by self-play and reinforcement learning are discussed in Section \ref{sec:conclusion} and will be the object of future work. 

However, despite great similarities with the setting of AlphaGO, our problem has features of its own, which are addressed in this section. First, in the game playing setting, every action is perpetrated by one player with the intention of winning the game (i.e. pursue the objective at hand). In contrast, historical actions in power networks may stem from various motivations, which include protecting the grid (our objective), but also include scheduled maintenance actions and various other maneuvers unrelated to our objective. Because of the lack of data annotation regarding the purpose of actions, we must perform sophisticated preprocessing to prepare data suitable for our machine learning modeling. 
Second, in a game setting the risk vs. reward trade-off does not have the same implications and level of gravity. In power network applications, much greater levels of care must be given to assessing potential adverse effects of proposed actions, possibly discarding those which may be curing a given problems while triggering one of several others.

Because of these distinguishing features of the problem, our methodology for ``imitations learning" is split in two steps, which are described in this section: (1) Data generation; (2) Learning.

\subsection{Dataset generation: Extracting relevant actions}
\label{seq:stime}

To train our models, which will imitate human dispatchers, we need a large dataset of pairs \{network state, action taken\}. We describe in this section the method we used to obtain such a dataset.

Our work builds upon a wealth of data recorded by RTE.
Every 5 minutes, the consistent state of the grid is archived. We have available data from  November \nth{1} 2011, to present times. For this study, we use data until 2016 August \nth{7}. 
For each grid state, we have access to all the injections (injections are complex numbers having active power positive or negative values and  reactive power values; they include both ``productions" and ``consumptions" or ``loads"). We also know the nodal topology of grid and the voltage (angle and magnitude) for every node of the network. Accurate simulators of the physical grid can compute other quantities, such as the flows on lines using standard models such as AC load-flow simulators.
This represents approximately $485~000$ snapshots of the  French grid: each snapshot being a modeling of the French Very High Voltage and High Voltage network counting more than $11~000$ lines, an average of around $6~400$ buses, and around $7~000$ loads for $3~000$ productions. 

One pitfall of the data is the lack of annotation of the actions. Changes in network topology cannot be only attributable to remedial actions taken by dispatchers to protect the grid. For example, we cannot distinguish between corrective actions performed in response to unplanned contingencies (e.g. a line struck by lightning) and periodical maneuvers to check if a breaker can still be open/closed. Therefore, to obtain data that is useful for training, we must perform a ``detective work" and extract from available data plausible remedial actions by analyzing which action, if \emph{not} performed, would have led to an adverse change in network security. 

Of two possible types of actions (re-dispatching and changes in topology), our main focus here is on topological actions. This stems from two main reasons. First, in the literature, some methods have already been developed to tackle the re-dispatching problem such method include OPF (Optimal power flow)\cite{dommel1968optimal} or SCOPF (Security Constrained Optimal Power Flow) where  \cite{capitanescu2016critical} present most recent advances in such area. 
Second, as we previously mentionned, TSOs like RTE are more interested in topological remedial actions because they are generally less costly. 

To isolate the relevant changes in network topology, which could correspond to dispatcher actions responding to a problem or anticipating a situation that may yield to a problem, we propose an algorithm inspired by counterfactual reasoning~\cite{pearl2009causality}: "What would have happened if a given topological change $\tau$ had \emph{not} occurred?" To do that, we use a combination of real data and grid simulation. We proceed in two steps for which pseudo-code is provided:
\begin{itemize} 
\item Algorithm \ref{alg:cont}: Considering two grid states $g_t$, and $g_{t+h}$ at times $t$ and $t+h$, we check the potential outcome of \emph{not} having performed a change in topology by freezing the network topology at $t$ while imposing the injections that were observed in real data at $t+h$. The power flows and security criterion $\mathbb{S}$ are re-calculated by simulation. Unsafe networks are detected when security violations occur, indicating that a topological change may have played the role of a preventive ``remedial action". 
\item Algorithm \ref{alg:par}: Changes in topology occurring between $t$ and $t+h$ may have been motivated by other reasons than preventing the network to go out of its security operation regime (for reason of maintenance, for example). We post-process the data by looking for a minimal subset of actions, which bring the network back to a safe operation mode. 
\end{itemize} 

\begin{algorithm}[h]
\caption{Algorithm for finding unsafe networks. Observed grid states are denoted by $g_t$. Counterfactual grid states are denoted by $\tilde{g}_{t,h}$.}
\label{alg:cont}
 \begin{algorithmic}[1]
 \renewcommand{\algorithmicrequire}{\textbf{Input:}}
 \renewcommand{\algorithmicensure}{\textbf{Output:}}
 \REQUIRE $\{g_t\}_{t_{\text{min}} \leq t \leq t_{\text{max}} }$, $h_{\text{max}}$, $\mathbb{S}$
 \ENSURE  $\{ (t,~ h, ~s, ~\tilde{g}_{t,h}) \}$
 \\ \textit{Initialisation} :
  \STATE $\text{res} \leftarrow \{\}$
  \\ \textit{Main loop} :
  \FOR {$t \in [t_{\text{min}},t_{\text{max}}]$ }
  \FOR {$h \in [0, h_{\text{max}}]$ }
  \STATE create grid $\tilde{g}_{t,h}$ with the same injections than $g_{t+h}$ and the same topology than $g_t$
  \STATE $S = \mathbb{S}(\tilde{g}_{t,h})$
  \IF {($S \neq \emptyset $)}
  \FOR {$s \in S$}
  \STATE $\text{res.append}((t,~h,~s,~\tilde{g}_{t,h}))$
  \ENDFOR
  \ENDIF 
  \ENDFOR
  \ENDFOR
 \RETURN $\text{res}$ 
 \end{algorithmic} 
\end{algorithm}

\begin{algorithm}[h]
\caption{Algorithm for extracting minimal remedial actions.}
\label{alg:par}
 \begin{algorithmic}[1]
 \renewcommand{\algorithmicrequire}{\textbf{Input:}}
 \renewcommand{\algorithmicensure}{\textbf{Output:}}
 \REQUIRE $\{g_t\}_{t_{\text{min}} \leq t \leq t_{\text{max}} }$, $\{ (t,~ h, ~s,~\tilde{g}_{t,h}) \}$,  $\mathbb{S}$
 \ENSURE  $\{ (s, \tau, \tilde{g})\}$
 \\ \textit{Initialisation} :
  \STATE $\text{res} \leftarrow \{\}$
  \\ \textit{Main loop} :
  \FOR {$t,h,s,\tilde{g} \in \{ (t,~ h, ~s,~\tilde{g}_{t,h}) \}$}
      \STATE Compute the topological changes between $g_t$ and $g_{t+h}$, assign it to $\Gamma$
      \FOR{ $\tau \in subset(\Gamma)$}
        \IF{not $s \in \mathbb{S}(\tilde{g} \odot \tau)$}
            \STATE $\text{res.append}((s, \tilde{g}, \tau))$
        \ENDIF
      \ENDFOR
  \ENDFOR
 \RETURN $\text{res}$ 
 \end{algorithmic} 
\end{algorithm}

The output of Algorithm \ref{alg:cont} is then a list of security criteria not met in a stressed network, and the corresponding time-stamps. The output of Algorithm \ref{alg:par} is a list of topological changes that can be applied as remedial actions. 

\subsection{Model training: Imitate human experts\label{seq:imitate}}
Now that we have a clean database with pairs of \{$X$=stressed state, $Y$=remedial actions\}
we can learn from it. 
The main idea is to use learning machines to quickly propose and/or evaluate actions by learning from what the human would have done facing the same situation. This is often called \textit{supervised learning}, or \textit{imitation learning}. 
For instance, we may provide our learning machine with an ensemble of variables $X$ (in our case an encoding of the security issue-s- $s$ and the grid $g$) and teach it to produce the response $Y=\tau$. 
One of the main difficulties we have to face is that of encoding information: the structure and state of a power grid, including representing security issues, and the actions.
We propose and study several methods of encoding, restraining ourselves to the French power grid of which we have in depth knowledge.
One of them consists in simply enumerating all the important variables, for example the productions, the loads, the flows on each line or the voltage magnitude and angles and encode them with an arbitrary ``barcode". This first approach main seem too crude, but has proved useful in combination of deep learning neural network architectures that we have explored in our machine learning analyses. This also demonstrates the robustness of deep learning techniques to arbitrary input representation and their capability of learning internal representation even from unpreprocessed data as shown in \cite{lecun1998gradient} for example. This is a important feature to achieve our goal: model grid data is a complex task.


As learning machines, several neural network architectures have been envisioned and will be compared. One of the most promising ones, for which we have initial results reported in the next section and that could serve as benchmark for most advance study, involves a deep neural network, which predicts power flows from injections and topologies, simply coded with their ``barcode”. The benefit of this network is to quickly be able to evaluate the security status of a proposed power network topology by calculating  $\mathbb{S}$ from the neural network output. Such evaluation of $\mathbb{S}$ using a neural network is orders of magnitude faster than running the RTE simulator Hades2 (typically 100x for moderate size neural networks used on moderate size powergrid). Today, this first module must be combined with another system, which produces candidate topologies, including topologies proposed by dispatchers and re-combinations. We presently have a dictionary of $3~000$ topologies corresponding to preventive ``remedial actions". We envision that this set of remedial actions could be enriched with the help of data generating models such as GANs~\cite{goodfellow2014generative} or can be ranked using a learning algorithm and then tested in real time with the preferred simulator of the dispatcher.


\section{Main results}
\label{sec:results}

The previous algorithm \ref{alg:cont} and \ref{alg:par} have been run through the first six months of 2012. To make the simulation tractable for a reasonable computer, some restriction have been imposed some restrictions: 
\begin{itemize}
\item[a)] we impose the window $h$ to be in the ensemble $\{$ 5 min, 10 min, 15 min, 30\text{min}, 45 min, 1h, 1h30, 2h, 2h30, 3h, 3h30, 4h, 4h30, 5h, 5h30, 6h, 7h, 8h, 9h, 10h, 11h, 12h, 23h, 23h30, 23h45, 24h $\}$ and not on the interval $[0, 24h]$ in algorithm \ref{alg:cont}. This restriction have been made in compliance with RTE experts, and preliminary results that showed a lot of redundancies.
\item[b)] we use a simplify version of the safety criterion use for operation support $\mathbb{S}$. The criterion used was that each line of the network must be bellow $95\%$ of its thermal limit. The operation safety criterion would lead to $10~000$ times more computation, as for each security assessment, a simulation retrieving each line one by one must be computed. \\
\item[c)] in algorithm \ref{alg:par}, only the subset of $\tau$ of cardinal one have been tested. This again is in compliance with the operators: it is quite rare that people need to act on different substations for security reason.
\end{itemize}

\begin{table}[H]
\centering
{\small
\begin{tabular}{c| c}
& Total \\
\hline
$\tilde{g}_{t,h}$ computed &  $1~163~940$ \\
$\tilde{g}_{t,h}$ unsafe & $81~476$\\
$\tilde{g}_{t,h}$ with one curative action & $17~587$\\
different curative actions & $3~266$ \\
lines stressed & $\mathbf{2~008}$ \\  
lines stressed with a curative action & $\mathbf{964}$\\
\end{tabular}
}
\caption{Results for obtained after launching algorithms \ref{alg:cont} and \ref{alg:par} on the six firsts months of 2012.}
\label{tab:res_par}
\end{table}

With this settings, the security around $1~250~000$ grids have been computed using our implementation of algorithm \ref{alg:cont} as shown in the table \ref{tab:res_par} (first row). This allowed us to identify more than $81~000$ stressed grids $\tilde{g}_{t,h}$ in an insecure state. On this $81~000$ insecure grids, we noted that $2~008$ lines have seen their flow exceed their thermal limit (fifth row). This represent around $18\%$ of the total number of lines present in the grid. This validation is compliant with expert knowledge of the French grid: there exists some part weaker than the others. \\

We can also note that in total, we have found $3~266$ unique remedial actions (two remedial actions are different if and only if they do not solve an overflow on the same line or if they do not act on the same substation, of if they do not change the nodal topology in the same manner). This means that, in the history, at least $3~266$ different topological actions could have been done for solving a security issue. \\

With these data collected, we intend to conduct a systematic comparison of learning machine architectures to propose and evaluate ``remedial actions". We have first started studying neural network architectures allowing us to evaluate ``remedial actions". As explained in the previous section, such learning machines take as input injections and topologies and predict power flows (which allow us to quickly calculate $\mathbb{S}$).

\begin{figure}[H]
    \centering
       \begin{tikzpicture}[scale=0.5,rotate=-90]
           \draw (0,0) -- ++(1,0)-- ++(0,1)-- ++(-1,0)-- ++(0,-1);
           \draw (0.5,0.5) node {{\small $p_p$} };
           \draw (2,0) -- ++(1,0)-- ++(0,1)-- ++(-1,0)-- ++(0,-1);
           \draw (2.5,0.5) node {{\small $p_v$} };
           \draw (4,0) -- ++(1,0)-- ++(0,1)-- ++(-1,0)-- ++(0,-1);
           \draw (4.5,0.5) node {{\small $c_p$} };
           \draw (6,0) -- ++(1,0)-- ++(0,1)-- ++(-1,0)-- ++(0,-1);
           \draw (6.5,0.5) node {{\small $c_q$} };
           \draw (8,0) -- ++(1,0)-- ++(0,1)-- ++(-1,0)-- ++(0,-1);
           \draw (8.5,0.5) node {{\small $c_q$} };
           \draw (10,0) -- ++(1,0)-- ++(0,1)-- ++(-1,0)-- ++(0,-1);
           \draw (10.5,0.5) node {{\small enc} };
           
           \draw[->] (0.5,1)--(2.5,3.45);
           \draw[->] (0.5,1)--(8.5,3.45);
           \draw[->] (2.5,1)--(2.5,3.45);
           \draw[->] (2.5,1)--(8.5,3.45);
           \draw[->] (4.5,1)--(2.5,3.45);
           \draw[->] (4.5,1)--(8.5,3.45);
           \draw[->] (6.5,1)--(2.5,3.45);
           \draw[->] (6.5,1)--(8.5,3.45);
           \draw[->] (8.5,1)--(2.5,3.45);
           \draw[->] (8.5,1)--(8.5,3.45);
           \draw[->] (10.5,1)--(2.5,3.45);
           \draw[->] (10.5,1)--(8.5,3.45);

           \draw (2.5,4) circle(0.55);
           \draw (5.5,4) node {{\small $\vdots$} };
           \draw (8.5,4) circle(0.55);
           
           \draw (5.5,5.5) node {{\small $\dots$} };
           
           
           \draw (2.5,7) circle(0.55);
           \draw (5.5,7) node {{\small $\vdots$} };
           \draw (8.5,7) circle(0.55);
           
           \draw (5.5,5.5) node {{\small $\dots$} };

           \draw[->] (2.5,7.55)--(2,10);
           \draw[->] (2.5,7.55)--(4,10);
           \draw[->] (2.5,7.55)--(6,10);
           \draw[->] (2.5,7.55)--(8,10);
           
           \draw[->] (8.5,7.55)--(2,10);
           \draw[->] (8.5,7.55)--(4,10);
           \draw[->] (8.5,7.55)--(6,10);
           \draw[->] (8.5,7.55)--(8,10);
           
           \draw (1.5,10)-- ++(1.5,0)-- ++(-0.75,1.5)-- ++(-0.75,-1.5);
           \draw (2.25,10.5) node {{\small $p_q$} };
           \draw (3.5,10)-- ++(1.5,0)-- ++(-0.75,1.5)-- ++(-0.75,-1.5);
           \draw (4.25,10.5) node {{\small $c_v$} };
           \draw (5.5,10)-- ++(1.5,0)-- ++(-0.75,1.5)-- ++(-0.75,-1.5);
           \draw (6.25,10.5) node {{\small $f_A$} };
           \draw (7.5,10)-- ++(1.5,0)-- ++(-0.75,1.5)-- ++(-0.75,-1.5);
           \draw (8.25,10.5) node {{\small $f_{MW}$} };
           

        \end{tikzpicture}
        
        \caption{Representation of the architecture of the artificial neural network used for the flows approximation. Squares represents input, triangles output and hidden unit are represented by circle. Each arrow represent a trainable parameter (weight). The network architecture is "fully connected": every unit of a given layer is connected to every unit of the following layer.}
        \label{fig:nn_arch}
    \end{figure}
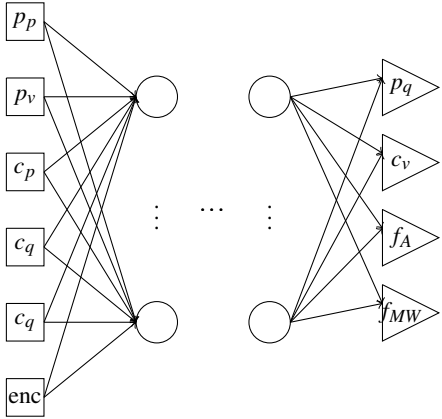

Our preliminary study includes testing artificial neural network for approximating load-flow of Matpower \cite{Zimmerman11matpowersteadystate} (test cases "case30" coming originally from \cite{alsac1974optimal} and "case118"\footnote{This test case "represents a portion of the American Electric Power System (in the Midwestern US) as of December, 1962". More information can be found at \href{http://www2.ee.washington.edu/research/pstca/pf118/pg_tca118bus.htm}{www2.ee.washington.edu/research/pstca/pf118}.}). Using these power grids we taught the neural networks to predict the outcome of a given outage. The neural networks are trained using the Tensorflow framework. An example of the architecture used to approximate the load-flow can be found in figure \ref{fig:nn_arch}. \\

This study has been conducted by first simulating a lot of \emph{plausible} grid state, making the injections (productions and consumptions) vary. The workflow to obtain such a database is the following:
\begin{enumerate}
\item Get the grid in the proper format used by Hades2
\item Disconnect one line.
\item Sample the active loads based on the 2012 French loads consumptions
\item Sample the reactive loads from the historical distribution $\dfrac{p}{q}$ calibrated on the French power grid
\item Sample active productions value from active loads value:
\begin{itemize}
\item Disconnect randomly some productions (to take into account the fact that not all productions are functioning at a given time)
\item dispatch the loads power according $p_\text{max}$
\item add noise
\end{itemize}
\item The voltages of the productions are not modified
\end{enumerate} 

Once the data base has been built, it has been divided in 3. One part (50 \%) for training the model, another one (25\%) for fitting the meta parameters, and at last a third one for testing and reporting results. The example for which the results are given are then never seen during any part of the training. \\

For each line disconnection, we ran $n_s = 10 000$ simulations with $10~000$ different productions / loads values. The matpower 30 buses grid count 41 lines, making in total $ \underbrace{n_s}_{\text{no line are disconnected}} + 41\times \underbrace{n_s}_{\text{one line is disconnected}}$. For this grid, the test set counts then $2~100~00$ samples. For the bigger $118$ buses grid, we simulate $n_s = 5~000$ sample per configuration, and there is $199$ lines, so the test set counts $450~000$ rows. \\

In this first experiment, we try to approximate a load-flow computation. So we feed a neural network with with an architecture presented in figure \ref{fig:nn_arch}: the active $c_p$ and reactive $c_q$ loads value, the active production value $p_p$ as well as their voltages setpoints $p_v$ as inputs. We also give in input which line have been disconnected using a one-hot encoding $enc$. And we ask the neural network to predict the rest of the variables: reactive productions values $p_q$, the voltages at the buses where each load is connected $c_v$ and the flows. For the flows we decided to make the network compute the active power flow $f_{MW}$, and the current power flow $f_a$. The reactive power flow is not computed. 
We note that we did not feed the network with the $p_{\text{min}}$ or $p_{\text{max}}$ values for the productions. One of the task of the neural network will be to balance the loads to take into account the losses for example.
To evaluate the performance of our models, we will use the Mean Absolute Error (MAE) and the Mean Absolute Percentage Error. If $y^{\text{true}}$ denotes the vector (of size $n$) of the true values, and $\hat{y}$ the vector of the predicted values (also of size $n$), we have:
\[ \left\{ 
\begin{aligned}
MAE(\hat{y},y^{\text{true}}) & \eqdef \frac{1}{n}.\sum_{i=1}^{n} \abs{\hat{y}_i - y^{\text{true}}_i} \\
MAPE(\hat{y},y^{\text{true}}) & \eqdef \frac{1}{n}.\sum_{i=1}^{n} \abs{ \frac{\hat{y}_i - y^{\text{true}}_i}{y^{\text{true}}_i} } 
\end{aligned}
\right.
\]

\begin{table}
\centering
{\small
\begin{tabular}{c|c|c}
Variable & 30 buses & 118 buses \\
&  MAE (MAPE) & MAE (MAPE) \\
\hline
$c_v$ (V) & 19 (0.02\%) & 190 (0.01\%)\\
$p_q$ (MVAr) &  1.75 (1.50\%) & 16.6 (1.81 \%) \\
$f_A$ (A) & 8.63 (1.23\%) & 62.1 (2.3 \%) \\
$f_{MW}$  (MW)& 0.7 (0.76\%) &  7.4 (1.12\%)\\
\end{tabular}
}
\caption{Mean absolution error (MAE) and mean relative percentage error (MAPE) for each of the output of the learning algorithm. The value are computed over the entire test set from data never seen during the training. The lines correspond to: $c_v$ the voltages of the bus where the load is connected, $p_q$ the reactive power produce by a plant, $f_A$  the current flow on a line, and $f_{MW}$ the active power flow on a line.}
\label{tab:res_nn}
\end{table}
As we can see from the table \ref{tab:res_nn}, the neural networks achieve great performance. They are able to predict the output of the load-flow with an error close to $1\%$ for the $30$ buses grid and around $2\%$ for the $118$ buses, which is enough for looking at curative actions, as one will see in the next section. We must note that no special care have been taken to feed the data in the neural network. Further studies will focus on the matter. We believe that this could greatly improve the performance. \\

To be complete, training the model for the 30 buses grid took 18h 31min on a computer with an high-end GPU (Nvidia GTX 1080) and 20h 03min (in this case, the error was still decreasing when at the time of writing). Once the model are trained, the computation of load-flow is very fast. Computing $5~000$ security analysis for the "N-1" criterion for the 30 buses grid ($210~000$ load-flows) took only $1.56s$ on an intel i5 2 cores laptop processor. For comparison, generating the dataset using a much faster i7 processor took $123.7$s. This lead to computation time speed-up of around $80$. Concerning the $118$ buses the speed-up is about $450$ ($1~432$s to generate the data and $3.01$s to compute $2~500$ security analysis, representing $450~000$ load-flows using the trained model).

The main drawback of this method consist in the fixed topology settings.Only lines disconnection are taken into account. It is for now impossible to perform more complex topological changes on the power grid. We are currently working on this issue, and preliminary results seems promising. Even with more complex topological changes, the error is around $2-3\%$ for the 30 buses grid. No experiment have been done concerning the 118 buses grid yet. Once such a model will be available, we will be able to run the algorithms \ref{alg:cont} and \ref{alg:par} with the standard "N-1" security criterion. This could also allow us to test more topological changes. In summary, drastically reducing the computation time could allow us to find more historical curative actions. 

After building such a data base of curative actions, the next step will be to learn to mimic the human. The encouraging performance of artificial neural network in various supervised learning setting. The first encouraging results concerning flows approximations made us optimistic regarding the possibility to predict, based on human decisions, the substation for which we topology must be changed for security issue. The training of this model will be made with the data obtain after running algorithm \ref{alg:par}. The remedial action from which the algorithm will learn will concern the unsafe grid $\hat{g}_{t,h}$. That's where the time window $h_{\text{max}}$ play an important role. The time interval must be long enough to capture some possible remedial action, but narrow enough such that the grid $\hat{g}_{t,h}$ is "realistic" (eg that this simulated grid state is "close" enough to a grid that could have happened in real time). That's why we did not compute all the grid in the interval $\{$ 5 mins, $\dots$, 24 hours $\}$: so grid where completely unrealistic. For example applying the injections plan of peak time over the grid topology that was in operation at lowest load level often results in divergence of the load-flow.


\section{Links with operational decision processes \label{seq:inte}}
In this section, we will explain our view about the possible usage of our method as a tool to help the real time operations. \\

Let's consider that we have at our disposal the models discussed in the previous section:
\begin{itemize}
\item[M1] which approximate a load-flow computation very rapidly
\item[M2] that is able, given a safety issue and a grid state to predict accurately on which substation we can act. 
\end{itemize}
First, as the grid evolve the Model 1 and Model 2 describe above could be learned from time to time, for example during the week-end, or if a greater computation power is available, during the night if time allows it. \\

Then the future real time operation framework could look like:
\begin{enumerate}
\item Use standard tools to assess whether or not a grid is secure. This could be done with standard computation, such as a load-flow computation and the "N-1" criterion. To speed-up the computation, and get faster results, one could also use model M1 to pre-screen the contingencies that will most probably cause at least one overload.
\item If there is some non secure contingency detected, one could then use model M2 to predict on which substation a topological action is worth looking for. Let's name $sub_i$ this substations.
\item After such substation is detected, we could enumerate all possible action doable at the time of the study in $sub_i$. We could rapidly assess if a possible curative action is found or not. 
\item Then are 2 cases:
\begin{itemize}
\item If a possible change has been found with this method, we will use accurate model such as load-flow as well as models that take into account dynamic phenomenon to check that the action found remove the security issue, and that it does not cause any problem elsewhere.
\item Or no action have been found. In this case, we let the operator the choice of which action to do. But we can tell him that it is most probably useless to seek for a topological action in the substation $sub_i$.
\end{itemize}
\end{enumerate}

As one can see, this framework offers a lot of flexibility. One can for example decide at step 3 to look at the $k$ "most likely" substations where a topological curative action can take place. This would of course increase the computation time, but it will be more likely to find one. Also, this method does allow for operators to take the control at any moment. For example, it is always possible to stop the research of curative actions, and the algorithm will be able to tell which actions have been (unsuccessfully) tested quite easily. \\

Most of all the security assessment can be performed after the action have been chosen by the machine, the fast approximation are only relevant for exploring the curative actions space. Another set of methods, including dynamic simulation of the changes in the grid will take place after the selection of the right action. The proposed method is then a mixture of different approaches. We use machine learning methods to search the curative action. The security check are performed with very well established method, relying on simulation of physical systems.




\section{Discussion and Conclusion}
\label{sec:conclusion}

This paper proposed to generate candidate remedial actions to dispatchers in order to maintain a power network in `a safe state, using machine learning techniques. With our method, remedial actions are rank-orders in order of increasing cost (costs for re-dispatching are typically much higher than costs for modifying network topology) and then tested with simulators used today by dispatchers before being proposed to them.
Our methodology requires first extracting from historical data actual actions that were performed and have been evaluated to have a positive influence (protect against possible network issues to be avoided, such as power flows exceeding lines thermal limits). That alone is a non-trivial problem because (1) many actions performed on the network are not protective actions (they may include maintenance actions and miscellaneous maneuvers); (2) there is no centralized and uniform record of why given actions are performed; 
(3) the consequences of not performing given actions are not observed, hence it is difficult to assess how effectively protective given actions may be.
We devised and implemented an algorithm based on the causal concept of counterfactuals, which allows us to identify actions that have had beneficial effects (or more precisely, without which the network would have incurred adverse effects). Such training data will be used to train learning machines in an supervised way to imitate the actions of dispatchers or to evaluate rapidly candidate actions. This allows us to generalize and generate ranked lists of remedial actions in situations never seen before. We tested our method on small well-known test cases and obtained promising preliminary results. 

The proposed methodology counts multiple advantages. The first one is to be able to explore a vast number of possible curative actions, thanks to the very fast approximation of flows. But most importantly, we think that this method will proposed realistic remedial actions thanks to learning from operators expertise by observation.
Or methodology is to some extent inspired by game playing machine learning programs such as AlphaGo of Google Deepmind.
\cite{silver_mastering_2016}.
Further work will consist in learning using \textit{reinforcement learning} to refine our learning machine. In the same manner than AlphaGo improved itself by self-play, after being only initially trained to imitate the play of famous Go players, we intend to use the RTE simulator to generate millions of new situations and let the learning machine propose candidate remedial solutions and learn from its errors to progressively improve (i.e. decrease cumulative costs). In combination with Monte Carlo Tree Search (as used by AlphaGo), we believe that this could be a powerful way of improving policy learning. 
Other avenues of research include seeking the worst case events that could happen after a remedial action took place, following the work of\cite{capitanescu2011day} and \cite{fliscounakis2013contingency}, for example. \\ 

Another possible extension would be to used the proposed framework in more generic settings in the context of mid- to long-term studies, where real-time actions must be taken into account (the GARPUR\footnote{GARPUR: \textit{Generally Accepted Reliability Principle with Uncertainty modelling and through probabilistic Risk assessment} (\href{http://www.garpur-project.eu/}{http://www.garpur-project.eu/}), is an European project which "aims to maintain power system performance at a desired level, while minimizing the socio-economic costs of keeping the power system at that performance level".} project would be an example), 
or for the classification of contingencies in the case of the I-TESLA project\footnote{I-TESLA stands for Innovative Tools for Electrical System Security within Large Areas. I-TESLA  is a European project (\href{http://www.itesla-project.eu/}{http://www.itesla-project.eu/}) aiming at ``improving network operations with a new security assessment tool".}. \\
We also intend to explore many remedial action recombination strategies to enrich the space of exploration, in the spirit of genetic algorithms. While our approach will initially draw on classical Markov Decision Processes, assuming largely quasi-total observability of the grid state and dispatcher actions, we will progressively incorporate more realism and complexity and devise methods having only partial knowledge of the overall situation, which may occur in case of delayed information transmission, and move into the realm of more complex models such as Partially Observable Markov Decision Processes (POMDP).

\bibliographystyle{IEEEtran}
\bibliography{references}




\end{document}